\documentclass{article} 
\usepackage{iclr2024_conference,times}

\usepackage{amsmath,amsfonts,bm}

\def\eqref#1{equation~\ref{#1}}

\def\1{\bm{1}}

\DeclareMathAlphabet{\mathsfit}{\encodingdefault}{\sfdefault}{m}{sl}
\SetMathAlphabet{\mathsfit}{bold}{\encodingdefault}{\sfdefault}{bx}{n}

\usepackage{hyperref}
\usepackage{url}
\usepackage{booktabs}
\usepackage{graphicx}
\usepackage{multirow}
\usepackage{placeins}

\title{Noise2Noise Denoising of CRISM \\ Hyperspectral Data}

\author{Robert Platt$^{1,2,3}$, Rossella Arcucci$^{1,3}$,  C\'edric M. John$^{4}$\\
$^{1}$Department of Earth Science and Engineering, Imperial College London \\
$^{2}$I-X, Imperial College London \\
$^{3}$Data Science Institute, Imperial College London \\
$^{4}$Digital Environment Research Institute (DERI), Queen Mary University of London \\
\texttt{\{rp1818, r.arcucci\}@ic.ac.uk}, \texttt{\{cedric.john\}@qmul.ac.uk} \\
}

\iclrfinalcopy 
\begin{document}

\maketitle

\begin{abstract}
Hyperspectral data acquired by the Compact Reconnaissance Imaging Spectrometer for Mars (CRISM) have allowed for unparalleled mapping of the surface mineralogy of Mars. Due to sensor degradation over time, a significant portion of the recently acquired data is considered unusable. Here a new data-driven model architecture, Noise2Noise4Mars (N2N4M), is introduced to remove noise from CRISM images. Our model is self-supervised and does not require zero-noise target data, making it well suited for use in Planetary Science applications where high quality labelled data is scarce. We demonstrate its strong performance on synthetic-noise data and CRISM images, and its impact on downstream classification performance, outperforming benchmark methods on most metrics. This allows for detailed analysis for critical sites of interest on the Martian surface, including proposed lander sites.
\end{abstract}

\section{Introduction}

The hyperspectral data collected by the Compact Reconnaissance Imaging Spectrometer for Mars (CRISM) on-board NASA's Mars Reconnaissance Orbiter (MRO) mission has allowed for the highest resolution spectral mapping of the Martian surface to date. During its operation from 2006 to 2022, CRISM captured over 33,000 targeted observations and mapped 86\% of the planetary surface~\citep{seelos_crism_2023}. CRISM data have allowed for detailed mapping of the surface mineralogy, and understanding of the geological evolution of the red planet~\citep{bishop_visible_2019}.

Despite the vast quantity of imagery acquired by CRISM, much of the data processing and image interpretation is accomplished manually. Significant technical expertise is required to interpret the imagery, so the reliance on manual efforts represents a large bottleneck in effective use of this data. In addition, CRISM data quality has decreased steadily since its launch, attributed to failing cryocooling systems~\citep{carter_hydrous_2013}. As a result, much of the more recent imagery ($>$2012) contains a sufficiently high noise fraction (see Fig.~\ref{low_high_noise_spectra}) to obscure absorption features, which restricts mineral identification and analysis~\citep{mandon_morphological_2021}. This limits our understanding of key sites of interest identified more recently, for example Oxia Planum, the proposed landing site for ESA's ExoMars rover~\citep{mandon_morphological_2021}. The scarcity of labelled or ground truth data presents an additional challenge, as even older CRISM data still contains noise from atmospheric and photometric effects. Therefore, noise-free ground truth data does not exist for Mars. This highlights the need for development of unsupervised or self-supervised approaches~\citep{fernandes_generation_2022, pletl_spectral_2023}. To that end, our work introduces a new, self-supervised deep learning architecture for denoising hyperspectral signals, with no requirement for paired training data. Additionally, our model expects target data which itself is noisy, which is critical for application to Planetary Sciences where perfect data is rarely available.   The code used to develop this work can be found here: \url{https://github.com/rob-platt/n2n4m}

\begin{figure}
    \centering
    \scalebox{0.70}{\includegraphics[]{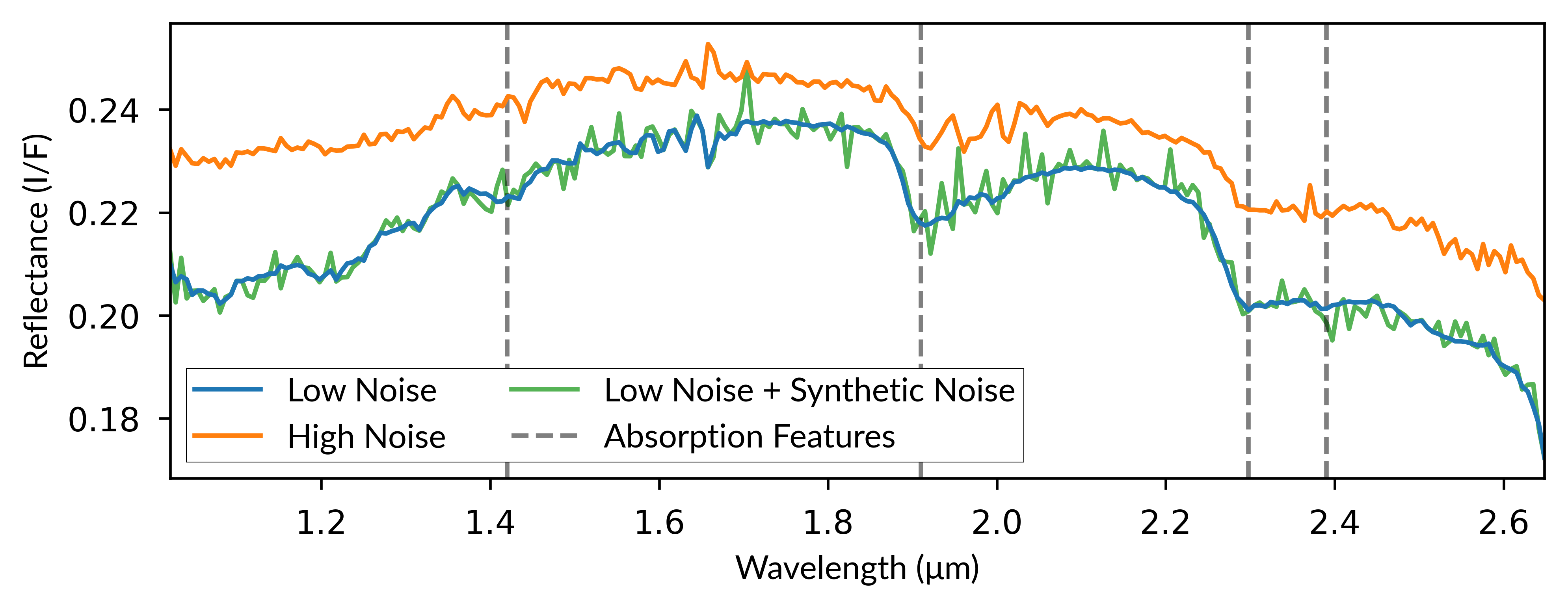}}
    \caption{Comparison of a low-noise and high-noise pixel spectra of the same location extracted from two CRISM images (FRT0000A053 and FRS000364CA) acquired 7 years apart. Absorption features to identify the surface material as an Mg/Fe-Smectite are highlighted. Shown in green is the low-noise spectra with synthetic Gaussian noise added.}
    \label{low_high_noise_spectra}
\end{figure}

\subsection{Related Works}

Two broad approaches have been developed to reduce noise and artefacts prevalent in CRISM data. The first uses and builds upon the so-called ``Volcano-Scan" correction, the current method applied in the preprocessing pipeline for CRISM IR images to correct for atmospheric effects~\citep{murchie_crism_2022, murchie_compact_2009, seelos_data_2011}. The CRISM Iterative Recognition and Removal of Unwanted Spiking (CIRRUS) and the Complement to CRISM Analysis Toolkit (CoTCAT) use statistical methods to reduce noise and build on ``Volcano-Scan" products~\citep{parente_new_2008, bultel_description_2015}. Similarly~\cite{carter_automated_2012} developed a tool (OCAT) using a Fast Fourier Transform (FFT) to generate a smoothed signal. 

A second approach developed around using radiative transfer modelling, utilizing a physical forward model (DISTORT) to simultaneously post-process and denoise the spectra~\citep{mcguire_mrocrism_2008}. Developments included application of Maximum Likelihood Estimation to the results~\citep{kreisch_regularization_2017}, and adding hypothesis testing for noise and data distribution~\citep{he_quantitative_2019}.  \cite{itoh_new_2020} developed a method to directly estimate the atmospheric transmission from the image itself, to simultaneously apply atmospheric correction and denoising. Both groups of methods have limitations - statistical methods can over-smooth key features \citep{bultel_description_2015}, whilst DISTORT-based methods have higher computational requirements and long ($\sim$2hrs per image) processing time \citep{itoh_new_2020}. Data-driven approaches remains unexplored.

A plethora of Machine Learning (ML) approaches have been developed for denoising of signal or image data, based on Convolutional Neural Networks (CNNs) \citep{tian_deep_2020, zhang_ffdnet_2018}, Autoencoders \citep{gondara_medical_2016}, and Diffusion models \citep{kawar_denoising_2022}. Of particular interest is the work of~\citep{lehtinen_noise2noise_2018}, who implemented a CNN where the target images are not assumed to be noise free. This is critical for application to CRISM data, where there is no noise-free ground truth.

\section{Methods}
\label{Methods}

The data used for this work is derived from the CRISM Machine Learning Toolkit dataset~\citep{plebani_machine_2022}. This consists of 586,599 pixels extracted from 77 low-noise CRISM images, associated with 39 corresponding mineral labels. The base Targeted Reduced Data Record (TRDR) images were processed using MarsSI~\citep{quantin-nataf_marssi_2018} to apply standard photometric and volcano-scan atmospheric corrections. Each IR spectra contains 350 channels between 1.0210\textmu m-2.6483\textmu m and 2.8070\textmu m-3.4769\textmu m, with channels known to contain artefacts excluded as in the original data \citep{plebani_machine_2022}. Additionally, 150,000 spectrally inert or ``bland", pixels identified by \cite{plebani_machine_2022} from a wider range of images were similarly processed and added to the dataset to better reflect the true class distribution. Synthetic noise is then added to the spectra to create high-noise training examples. The model objective is to reconstruct the low-noise signal from the high-noise signal as input.

\subsection{Preprocessing}

Two further preprocessing steps were applied to clean the data - imputation of bad values (I/F $>$ 1) and of data where significant atmospheric artefact remained between 1.91\textmu m and 2.08\textmu m due to issues with the volcano-scan correction. The data was split into train, test, and validation sets by image to ensure generalisation to unseen imagery. In addition, the Serpentine and Muscovite mineral classes were restricted to the test set only, to provide insight to the models ability to generalise to unseen mineral classes.

Synthetic noise was added to the low noise data to create training data (Fig. \ref{low_high_noise_spectra}). The synthetic noise partially obscures some of the key absorption features (e.g. the absorption at 1.42$\mu$m in Fig. \ref{low_high_noise_spectra}) similarly to real high noise spectra. This synthetic noise was created by sampling from a Gaussian distribution with zero mean, following \cite{bultel_description_2015}. An additional, uniformly sampled random variable is added to increase the inter-channel variability, which better reflects the observed noise distribution \citep{bultel_description_2015}. 

\subsection{Model and Training Scheme}

Our approach implemented a 1-Dimensional convolutional neural network (1D-CNN) in a U-Net architecture. This was inspired by the Noise2Noise (N2N) denoising work \citep{lehtinen_noise2noise_2018}, which showed that when using a U-Net architecture, a model can be trained effectively to denoise natural images despite the presence of noise in the target data. Our new model architecture, Noise2Noise for Mars (N2N4M), introduces important architectural changes to fit hyperspectral data. Instead of 2-dimensional convolution, we use 1-dimensional convolution layers to leverage features in the rich spectral dimension unique to hyperspectral imagery, rather than spatial data. Transposed convolution layers were used rather than nearest-neighbour upsampling in the decoder, and kernels of size 5 were found to be the most performant during hyperparamter tuning. N2N4M was trained for 100 epochs, and the validation dataset was used to tune learning rate, batch size, and model architecture. The final model architecture and exact training parameters can be found in Appendix \ref{appendix}. This architecture allows us to apply the model to CRISM data, where no zero-noise ground truth data exists, which has restricted data-driven approaches to the problem. 

\begin{figure}[h]
    \centering
    \includegraphics[width=\textwidth]{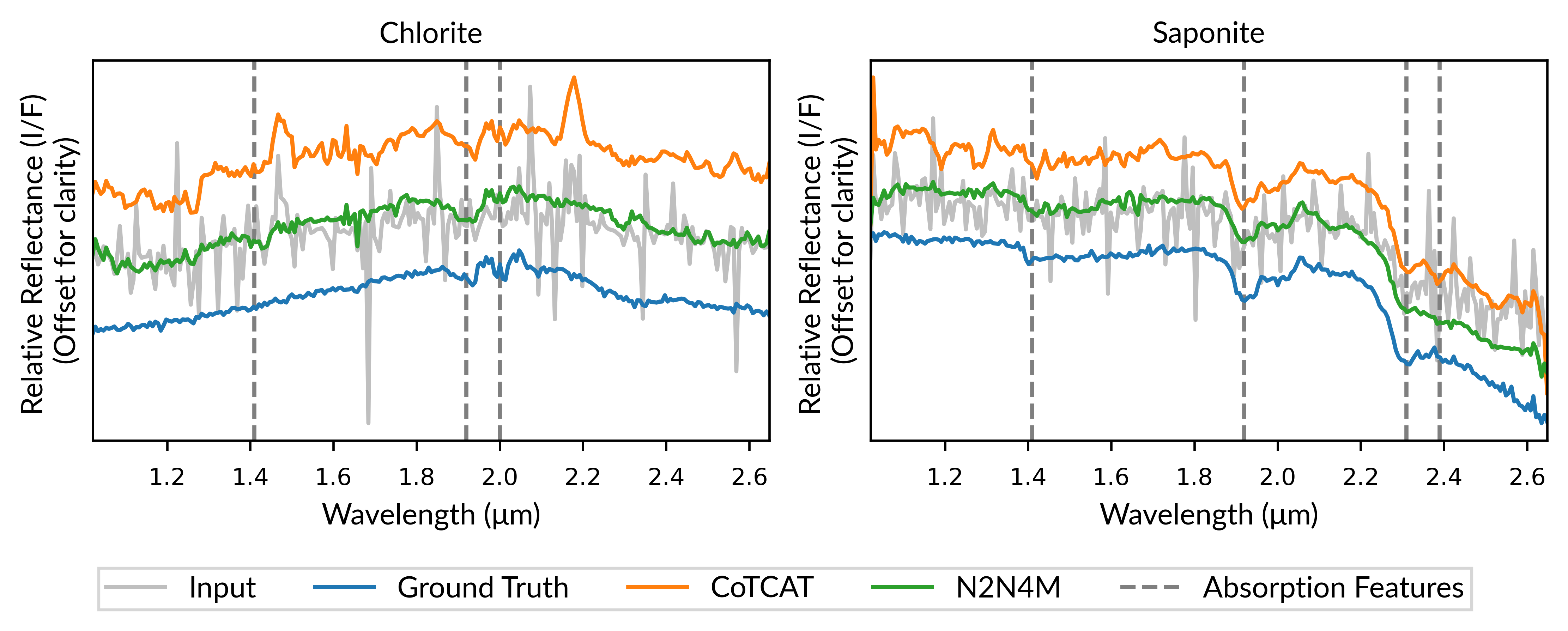}
    \caption{Qualitative comparsion of results on unseen spectra from two different mineral classes. Shown are the high-noise input, low-noise target, and high-noise data denoised by CoTCAT, and N2N4M (ours). Absorption features for mineral identification are highlighted.}
    \label{synthetic_results}
\end{figure}

\section{Results}

Model performance was tested with three approaches; denoising ability on synthetic noise data, downstream classification performance, and application to real imagery. The combination of these perceptual and quantitative metrics allow for a rigorous evaluation of the ability and limitations of the model. The Complement to CRISM Analysis Toolkit (CoTCAT) model \citep{bultel_description_2015}, and Savitzky-Golay filtering \citep{gorry_general_1990} were used as benchmarks for comparison.

\begin{figure}
    \centering
    \includegraphics[width=\textwidth]{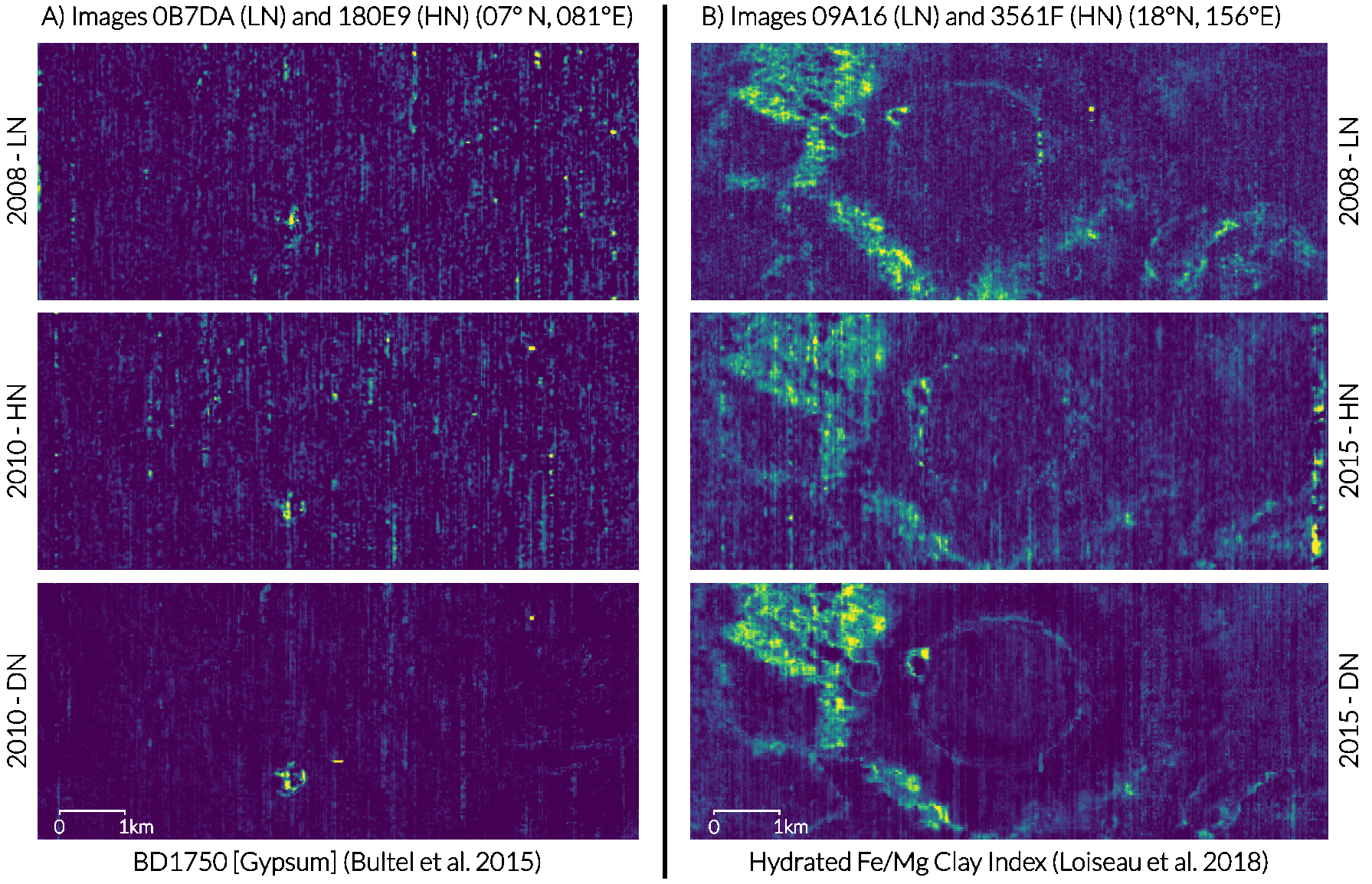}
    \caption{Two pairs of test images, from top to bottom: low-noise reference image (LN), high-noise image (HN), denoised image (DN). Images shown as the strength of absorption features (listed at base) in each pixel, which highlight outcrops of hydrated minerals identified in the reference image.}
    \label{real_image_results}
\end{figure}

When denoising unseen synthetic high-noise data, our model shows the lowest reconstruction error compared to the ground truth (Table \ref{quant_results}). A visual inspection of the results (see Fig.~\ref{synthetic_results} and Fig.~\ref{real_image_results}) shows that N2N4M removes more noise than CoTCAT whilst retaining the absorption features evident in the ground truth which are critical for mineral identification. For downstream classification, the Hierarchical Bayesian Model (HBM) developed by~\citep{plebani_machine_2022} was used, trained on low-noise data. The HBM predicts a mineral class label for an inputted spectra, with ground truth labels included in the original dataset by~\citep{plebani_machine_2022}. The results (Table \ref{quant_results}) are given as scores relative to performance on low-noise ground truth data, to highlight the impact of different denoising methods. N2N4M denoising results in a significant increase in most metrics over the benchmarks.

\begin{table}
\centering
\resizebox{\linewidth}{!}{%
\begin{tabular}{c|c|cccc} 
\hline\hline
\multirow{2}{*}{\textbf{Data}} & \textbf{Denoising Task} & \multicolumn{4}{c}{\textbf{Downstream Classification Task}}                                                       \\ 
\cline{2-6}
                               & \textbf{MSE}            & \textbf{Relative Accuracy} & \textbf{Relative F1 Score} & \textbf{Relative Precision} & \textbf{Relative Recall}  \\ 
\hline
Ground Truth~                  & N/A                     & 1.00                          & 1.00                          & 1.00                           & 1.00                         \\
Savitzky-Golay Filter          & $2.8\times 10^{-5}$                     & 0.01                       & 0.00                       & 0.42                        & 0.09                     \\
CoTCAT (Bultel et al. 2015)    & $5.0\times 10^{-6}$                     & 0.35                       & 0.43                       & \textbf{0.50}               & 0.41                      \\ 
\hline
N2N4M (Ours)                   & $\mathbf{4.7\times 10^{-6}}$            & \textbf{0.52}              & \textbf{0.48}              & \textbf{0.50}                        & \textbf{0.64}            
\end{tabular}
}
\caption{Benchmarking results for denoising, and the downstream classification task which uses the HBM from \cite{plebani_machine_2022}. Classification metrics are given relative to classifier performance on high quality, low-noise ground truth data. Best performance is highlighted in bold.}
\label{quant_results}
\end{table}

Image evaluation was performed on two CRISM image pairs. In each case, a low-noise, high quality image and high-noise, low quality image had been acquired over the same spatial location. The high-noise image in each pair was denoised using N2N4M, before standard post-processing steps were applied to all images using the ratioing method described by~\cite{plebani_machine_2022}. Summary parameters are a widespread method to visualise specific mineral suites in hyperspectral images~\citep{viviano_revised_2014} and allow for a visual assessment of denoising quality on real data. Summary parameters for known outcrops previously identified in the high quality imagery \citep{bultel_description_2015, mandon_morphological_2021} were calculated for all results (Fig. \ref{real_image_results}). The denoised images show a clear increase in perceptual quality over the original, with mineral outcrops highlighted with greater definition and clearer spatial extents. They compare favourably with the reference images and do not exhibit any spurious outcrops.

\section{Discussion and Future Works}

Our proposed model shows promising results on both synthetic and real noise hyperspectral CRISM imagery. The downstream use of this data is classification of surface mineralogy, and N2N4M denoised data shows a significant improvement over the benchmark in enabling this task.  As such automated classifiers are used to identify outcrop shapes and extents over exact mineral species identification \citep{plebani_machine_2022}, the 0.23 relative increase in recall by N2N4M over the benchmark represents a large increase in capabilities. When real imagery is assessed, there is a clear increase in visual quality post-denoising. Additionally, no artificial mineral outcrops are introduced which is critical for trust in the wider use of this model outside of examples where low noise imagery is available. Whilst N2N4M shows a clear step forward in performance, further development is needed for truly reliable denoising and classification. Many higher noise CRISM images display striping noise \citep{carter_automated_2012}, and methods to include this spatial component directly in the model should be explored.

\section{Conclusion}
A novel, data-driven method for denoising hyperspectral imagery of Mars where true no-noise ground truth is unavailable has been proposed. This approach is shown to be effective at removing noise whilst retaining key absorption features of spectra, for both synthetic noise and real example imagery. Our method is not restricted to the particular characteristics of SWIR spectra and could easily be applied in other domains e.g. Synthetic Aperture Radar (SAR), or Radio signals. Our work should allow for further analysis of much of the Martian surface mineralogy, and highlights the use-case for self-supervised ML methods in Planetary Sciences.


\bibliography{iclr2024_conference}

\begin{thebibliography}{25}
\providecommand{\natexlab}[1]{#1}
\providecommand{\url}[1]{\texttt{#1}}
\expandafter\ifx\csname urlstyle\endcsname\relax
  \providecommand{\doi}[1]{doi: #1}\else
  \providecommand{\doi}{doi: \begingroup \urlstyle{rm}\Url}\fi

\bibitem[Bultel et~al.(2015)Bultel, Quantin, and Lozac’h]{bultel_description_2015}
Benjamin Bultel, Cathy Quantin, and Loïc Lozac’h.
\newblock Description of {CoTCAT} ({Complement} to {CRISM} {Analysis} {Toolkit}).
\newblock \emph{IEEE Journal of Selected Topics in Applied Earth Observations and Remote Sensing}, 8\penalty0 (6):\penalty0 3039--3049, June 2015.
\newblock ISSN 2151-1535.
\newblock \doi{10.1109/JSTARS.2015.2405095}.

\bibitem[Carter et~al.(2012)Carter, Poulet, Murchie, and Bibring]{carter_automated_2012}
J.~Carter, F.~Poulet, S.~Murchie, and J.~P. Bibring.
\newblock Automated processing of planetary hyperspectral datasets for the extraction of weak mineral signatures and applications to {CRISM} observations of hydrated silicates on {Mars}.
\newblock \emph{Planetary and Space Science}, 76:\penalty0 53--67, December 2012.
\newblock ISSN 0032-0633.
\newblock \doi{10.1016/j.pss.2012.11.007}.
\newblock URL \url{https://www.sciencedirect.com/science/article/pii/S0032063312003625}.

\bibitem[Carter et~al.(2013)Carter, Poulet, Bibring, Mangold, and Murchie]{carter_hydrous_2013}
J.~Carter, F.~Poulet, J.-P. Bibring, N.~Mangold, and S.~Murchie.
\newblock Hydrous minerals on {Mars} as seen by the {CRISM} and {OMEGA} imaging spectrometers: {Updated} global view.
\newblock \emph{Journal of Geophysical Research: Planets}, 118\penalty0 (4):\penalty0 831--858, April 2013.
\newblock ISSN 2169-9100.
\newblock \doi{10.1029/2012JE004145}.
\newblock URL \url{https://onlinelibrary.wiley.com/doi/abs/10.1029/2012JE004145}.
\newblock \_eprint: https://onlinelibrary.wiley.com/doi/pdf/10.1029/2012JE004145.

\bibitem[Fernandes et~al.(2022)Fernandes, Pletl, Thomas, Rossi, and Elser]{fernandes_generation_2022}
Michael Fernandes, Alexander Pletl, Nicolas Thomas, Angelo~Pio Rossi, and Benedikt Elser.
\newblock Generation and {Optimization} of {Spectral} {Cluster} {Maps} to {Enable} {Data} {Fusion} of {CaSSIS} and {CRISM} {Datasets}.
\newblock \emph{Remote Sensing}, 14\penalty0 (11):\penalty0 2524, January 2022.
\newblock ISSN 2072-4292.
\newblock \doi{10.3390/rs14112524}.
\newblock URL \url{https://www.mdpi.com/2072-4292/14/11/2524}.
\newblock Number: 11 Publisher: Multidisciplinary Digital Publishing Institute.

\bibitem[Gondara(2016)]{gondara_medical_2016}
Lovedeep Gondara.
\newblock Medical {Image} {Denoising} {Using} {Convolutional} {Denoising} {Autoencoders}.
\newblock In \emph{2016 {IEEE} 16th {International} {Conference} on {Data} {Mining} {Workshops} ({ICDMW})}, pp.\  241--246, December 2016.
\newblock \doi{10.1109/ICDMW.2016.0041}.
\newblock URL \url{https://ieeexplore.ieee.org/abstract/document/7836672}.
\newblock ISSN: 2375-9259.

\bibitem[Gorry(1990)]{gorry_general_1990}
Peter~A. Gorry.
\newblock General least-squares smoothing and differentiation by the convolution ({Savitzky}-{Golay}) method.
\newblock \emph{Analytical Chemistry}, 62\penalty0 (6):\penalty0 570--573, March 1990.
\newblock ISSN 0003-2700.
\newblock \doi{10.1021/ac00205a007}.
\newblock URL \url{https://doi.org/10.1021/ac00205a007}.
\newblock Publisher: American Chemical Society.

\bibitem[He et~al.(2019)He, O’Sullivan, Politte, Powell, and Arvidson]{he_quantitative_2019}
Linyun He, Joseph~A. O’Sullivan, Daniel~V. Politte, Kathryn~E. Powell, and Raymond~E. Arvidson.
\newblock Quantitative {Reconstruction} and {Denoising} {Method} {HyBER} for {Hyperspectral} {Image} {Data} and {Its} {Application} to {CRISM}.
\newblock \emph{IEEE Journal of Selected Topics in Applied Earth Observations and Remote Sensing}, 12\penalty0 (4):\penalty0 1219--1230, April 2019.
\newblock ISSN 2151-1535.
\newblock \doi{10.1109/JSTARS.2019.2900644}.
\newblock URL \url{https://ieeexplore.ieee.org/document/8672185}.

\bibitem[Itoh \& Parente(2020)Itoh and Parente]{itoh_new_2020}
Yuki Itoh and Mario Parente.
\newblock A new method for atmospheric correction and de-noising of {CRISM} hyperspectral data.
\newblock \emph{Icarus}, 354:\penalty0 114024, July 2020.
\newblock ISSN 0019-1035.
\newblock \doi{10.1016/j.icarus.2020.114024}.
\newblock URL \url{https://www.sciencedirect.com/science/article/pii/S0019103520303857}.

\bibitem[Kawar et~al.(2022)Kawar, Elad, Ermon, and Song]{kawar_denoising_2022}
Bahjat Kawar, Michael Elad, Stefano Ermon, and Jiaming Song.
\newblock Denoising {Diffusion} {Restoration} {Models}.
\newblock \emph{Advances in Neural Information Processing Systems}, 35:\penalty0 23593--23606, December 2022.
\newblock URL \url{https://proceedings.neurips.cc/paper_files/paper/2022/hash/95504595b6169131b6ed6cd72eb05616-Abstract-Conference.html}.

\bibitem[Kreisch et~al.(2017)Kreisch, O'Sullivan, Arvidson, Politte, He, Stein, Finkel, Guinness, Wolff, and Lapôtre]{kreisch_regularization_2017}
C.~D. Kreisch, J.~A. O'Sullivan, R.~E. Arvidson, D.~V. Politte, L.~He, N.~T. Stein, J.~Finkel, E.~A. Guinness, M.~J. Wolff, and M.~G.~A. Lapôtre.
\newblock Regularization of {Mars} {Reconnaissance} {Orbiter} {CRISM} along‐track oversampled hyperspectral imaging observations of {Mars}.
\newblock \emph{Icarus}, 282:\penalty0 136--151, January 2017.
\newblock ISSN 0019-1035.
\newblock \doi{10.1016/j.icarus.2016.09.033}.
\newblock URL \url{https://www.sciencedirect.com/science/article/pii/S0019103516301579}.

\bibitem[Lehtinen et~al.(2018)Lehtinen, Munkberg, Hasselgren, Laine, Karras, Aittala, and Aila]{lehtinen_noise2noise_2018}
Jaakko Lehtinen, Jacob Munkberg, Jon Hasselgren, Samuli Laine, Tero Karras, Miika Aittala, and Timo Aila.
\newblock {Noise2Noise}: {Learning} {Image} {Restoration} without {Clean} {Data}, October 2018.
\newblock URL \url{http://arxiv.org/abs/1803.04189}.
\newblock arXiv:1803.04189 [cs, stat].

\bibitem[Mandon et~al.(2021)Mandon, Parkes~Bowen, Quantin-Nataf, Bridges, Carter, Pan, Beck, Dehouck, Volat, Thomas, Cremonese, Tornabene, and Thollot]{mandon_morphological_2021}
Lucia Mandon, Adam Parkes~Bowen, Cathy Quantin-Nataf, John~C. Bridges, John Carter, Lu~Pan, Pierre Beck, Erwin Dehouck, Matthieu Volat, Nicolas Thomas, Gabriele Cremonese, Livio~Leonardo Tornabene, and Patrick Thollot.
\newblock Morphological and {Spectral} {Diversity} of the {Clay}-{Bearing} {Unit} at the {ExoMars} {Landing} {Site} {Oxia} {Planum}.
\newblock \emph{Astrobiology}, 21\penalty0 (4):\penalty0 464--480, April 2021.
\newblock ISSN 1531-1074.
\newblock \doi{10.1089/ast.2020.2292}.
\newblock URL \url{https://www.liebertpub.com/doi/10.1089/ast.2020.2292}.
\newblock Publisher: Mary Ann Liebert, Inc., publishers.

\bibitem[McGuire et~al.(2008)McGuire, Wolff, Smith, Arvidson, Murchie, Clancy, Roush, Cull, Lichtenberg, Wiseman, Green, Marti, Milliken, Cavender, Humm, Seelos, Seelos, Taylor, Ehlmann, Mustard, Pelkey, Titus, Hash, and Malaret]{mcguire_mrocrism_2008}
Patrick~C. McGuire, Michael~J. Wolff, Michael~D. Smith, Raymond~E. Arvidson, Scott~L. Murchie, R.~Todd Clancy, Ted~L. Roush, Selby~C. Cull, Kim~A. Lichtenberg, Sandra~M. Wiseman, Robert~O. Green, Terry~Z. Marti, Ralph~E. Milliken, Peter~J. Cavender, David~C. Humm, Frank~P. Seelos, Kim~D. Seelos, Howard~W. Taylor, Bethany~L. Ehlmann, John~F. Mustard, Shannon~M. Pelkey, Timothy~N. Titus, Christopher~D. Hash, and Erick~R. Malaret.
\newblock {MRO}/{CRISM} {Retrieval} of {Surface} {Lambert} {Albedos} for {Multispectral} {Mapping} of {Mars} {With} {DISORT}-{Based} {Radiative} {Transfer} {Modeling}: {Phase} 1—{Using} {Historical} {Climatology} for {Temperatures}, {Aerosol} {Optical} {Depths}, and {Atmospheric} {Pressures}.
\newblock \emph{IEEE Transactions on Geoscience and Remote Sensing}, 46\penalty0 (12):\penalty0 4020--4040, December 2008.
\newblock ISSN 0196-2892.
\newblock \doi{10.1109/TGRS.2008.2000631}.
\newblock URL \url{http://ieeexplore.ieee.org/document/4685531/}.

\bibitem[Murchie et~al.(2022)Murchie, Guinness, and Slavney]{murchie_crism_2022}
Scott Murchie, Edward~A. Guinness, and Susan Slavney.
\newblock {CRISM} {Data} {Product} {Software} {Interface} {Specification} {Version} 1.3.7.6., June 2022.

\bibitem[Murchie et~al.(2009)Murchie, Seelos, Hash, Humm, Malaret, McGovern, Choo, Seelos, Buczkowski, Morgan, Barnouin-Jha, Nair, Taylor, Patterson, Harvel, Mustard, Arvidson, McGuire, Smith, Wolff, Titus, Bibring, and Poulet]{murchie_compact_2009}
Scott~L. Murchie, Frank~P. Seelos, Christopher~D. Hash, David~C. Humm, Erick Malaret, J.~Andrew McGovern, Teck~H. Choo, Kimberly~D. Seelos, Debra~L. Buczkowski, M.~Frank Morgan, Olivier~S. Barnouin-Jha, Hari Nair, Howard~W. Taylor, Gerald~W. Patterson, Christopher~A. Harvel, John~F. Mustard, Raymond~E. Arvidson, Patrick McGuire, Michael~D. Smith, Michael~J. Wolff, Timothy~N. Titus, Jean-Pierre Bibring, and Francois Poulet.
\newblock Compact {Reconnaissance} {Imaging} {Spectrometer} for {Mars} investigation and data set from the {Mars} {Reconnaissance} {Orbiter}'s primary science phase.
\newblock \emph{Journal of Geophysical Research: Planets}, 114\penalty0 (E2), 2009.
\newblock ISSN 2156-2202.
\newblock \doi{10.1029/2009JE003344}.
\newblock URL \url{https://onlinelibrary.wiley.com/doi/abs/10.1029/2009JE003344}.
\newblock \_eprint: https://onlinelibrary.wiley.com/doi/pdf/10.1029/2009JE003344.

\bibitem[Murchie et~al.(2019)Murchie, Bibring, Arvidson, Bishop, Carter, Ehlmann, Langevin, Mustard, Poulet, Riu, Seelos, and Viviano]{bishop_visible_2019}
Scott~L. Murchie, Jean-Pierre Bibring, Raymond~E. Arvidson, Janice~L. Bishop, John Carter, Bethany~L. Ehlmann, Yves Langevin, John~F. Mustard, Francois Poulet, Lucie Riu, Kimberly~D. Seelos, and Christina~E. Viviano.
\newblock Visible to {Short}-{Wave} {Infrared} {Spectral} {Analyses} of {Mars} from {Orbit} {Using} {CRISM} and {OMEGA}.
\newblock In Janice~L. Bishop, James~F. Bell~III, and Jeffrey~E. Moersch (eds.), \emph{Remote {Compositional} {Analysis}}, pp.\  453--483. Cambridge University Press, 1 edition, November 2019.
\newblock ISBN 978-1-316-88887-2 978-1-107-18620-0.
\newblock \doi{10.1017/9781316888872.025}.
\newblock URL \url{https://www.cambridge.org/core/product/identifier/9781316888872%23CN-bp-23/type/book_part}.

\bibitem[Parente(2008)]{parente_new_2008}
Mario Parente.
\newblock A {New} {Approach} {To} {Denoise} {CRISM} {Images}.
\newblock In \emph{39th {Lunar} and {Planetary} {Science} {Conference}}, March 2008.
\newblock URL \url{https://www.lpi.usra.edu/meetings/lpsc2008/pdf/2528.pdf}.

\bibitem[Plebani et~al.(2022)Plebani, Ehlmann, Leask, Fox, and Dundar]{plebani_machine_2022}
Emanuele Plebani, Bethany~L. Ehlmann, Ellen~K. Leask, Valerie~K. Fox, and M.~Murat Dundar.
\newblock A machine learning toolkit for {CRISM} image analysis.
\newblock \emph{Icarus}, 376:\penalty0 114849, April 2022.
\newblock ISSN 00191035.
\newblock \doi{10.1016/j.icarus.2021.114849}.
\newblock URL \url{https://linkinghub.elsevier.com/retrieve/pii/S0019103521004905}.

\bibitem[Pletl et~al.(2023)Pletl, Fernandes, Thomas, Rossi, and Elser]{pletl_spectral_2023}
Alexander Pletl, Michael Fernandes, Nicolas Thomas, Angelo~Pio Rossi, and Benedikt Elser.
\newblock Spectral {Clustering} of {CRISM} {Datasets} in {Jezero} {Crater} {Using} {UMAP} and k-{Means}.
\newblock \emph{Remote Sensing}, 15\penalty0 (4):\penalty0 939, January 2023.
\newblock ISSN 2072-4292.
\newblock \doi{10.3390/rs15040939}.
\newblock URL \url{https://www.mdpi.com/2072-4292/15/4/939}.
\newblock Number: 4 Publisher: Multidisciplinary Digital Publishing Institute.

\bibitem[Quantin-Nataf et~al.(2018)Quantin-Nataf, Lozac'h, Thollot, Loizeau, Bultel, Fernando, Allemand, Dubuffet, Poulet, Ody, Clenet, Leyrat, and Harrisson]{quantin-nataf_marssi_2018}
C.~Quantin-Nataf, L.~Lozac'h, P.~Thollot, D.~Loizeau, B.~Bultel, J.~Fernando, P.~Allemand, F.~Dubuffet, F.~Poulet, A.~Ody, H.~Clenet, C.~Leyrat, and S.~Harrisson.
\newblock {MarsSI}: {Martian} surface data processing information system.
\newblock \emph{Planetary and Space Science}, 150:\penalty0 157--170, January 2018.
\newblock ISSN 0032-0633.
\newblock \doi{10.1016/j.pss.2017.09.014}.
\newblock URL \url{https://www.sciencedirect.com/science/article/pii/S0032063316304718}.

\bibitem[Seelos et~al.(2011)Seelos, Murchie, Humm, Barnouin, and Mor]{seelos_data_2011}
F~P Seelos, S~L Murchie, D~C Humm, O~S Barnouin, and F~Mor.
\newblock Data {Processing} and {Analysis} {Products} {Update} - {Calibration}, {Correction}, and {Visualisation}.
\newblock In \emph{42nd {Lunar} and {Planetary} {Sciences} {Conference}}, 2011.

\bibitem[Seelos et~al.(2023)Seelos, Seelos, Murchie, Novak, Hash, Morgan, Arvidson, Aiello, Bibring, Bishop, Boldt, Boyd, Buczkowski, Chen, Clancy, Ehlmann, Frizzell, Hancock, Hayes, Heffernan, Humm, Itoh, Ju, Kochte, Malaret, McGovern, McGuire, Mehta, Moreland, Mustard, Nair, Núñez, O'Sullivan, Packer, Poffenbarger, Poulet, Romeo, Santo, Smith, Stephens, Toigo, Viviano, and Wolff]{seelos_crism_2023}
Frank~P. Seelos, Kimberly~D. Seelos, Scott~L. Murchie, M.~Alexandra~Matiella Novak, Christopher~D. Hash, M.~Frank Morgan, Raymond~E. Arvidson, John Aiello, Jean-Pierre Bibring, Janice~L. Bishop, John~D. Boldt, Ariana~R. Boyd, Debra~L. Buczkowski, Patrick~Y. Chen, R.~Todd Clancy, Bethany~L. Ehlmann, Katelyn Frizzell, Katie~M. Hancock, John~R. Hayes, Kevin~J. Heffernan, David~C. Humm, Yuki Itoh, Maggie Ju, Mark~C. Kochte, Erick Malaret, J.~Andrew McGovern, Patrick McGuire, Nishant~L. Mehta, Eleanor~L. Moreland, John~F. Mustard, A.~Hari Nair, Jorge~I. Núñez, Joseph~A. O'Sullivan, Liam~L. Packer, Ryan~T. Poffenbarger, Francois Poulet, Giuseppe Romeo, Andrew~G. Santo, Michael~D. Smith, David~C. Stephens, Anthony~D. Toigo, Christina~E. Viviano, and Michael~J. Wolff.
\newblock The {CRISM} investigation in {Mars} orbit: {Overview}, history, and delivered data products.
\newblock \emph{Icarus}, pp.\  115612, May 2023.
\newblock ISSN 0019-1035.
\newblock \doi{10.1016/j.icarus.2023.115612}.
\newblock URL \url{https://www.sciencedirect.com/science/article/pii/S0019103523001896}.

\bibitem[Tian et~al.(2020)Tian, Fei, Zheng, Xu, Zuo, and Lin]{tian_deep_2020}
Chunwei Tian, Lunke Fei, Wenxian Zheng, Yong Xu, Wangmeng Zuo, and Chia-Wen Lin.
\newblock Deep learning on image denoising: {An} overview.
\newblock \emph{Neural Networks}, 131:\penalty0 251--275, November 2020.
\newblock ISSN 0893-6080.
\newblock \doi{10.1016/j.neunet.2020.07.025}.
\newblock URL \url{https://www.sciencedirect.com/science/article/pii/S0893608020302665}.

\bibitem[Viviano-Beck et~al.(2014)Viviano-Beck, Seelos, Murchie, Kahn, Seelos, Taylor, Taylor, Ehlmann, Wiseman, Mustard, and Morgan]{viviano_revised_2014}
Christina~E. Viviano-Beck, Frank~P. Seelos, Scott~L. Murchie, Eliezer~G. Kahn, Kimberley~D. Seelos, Howard~W. Taylor, Kelly Taylor, Bethany~L. Ehlmann, Sandra~M. Wiseman, John~F. Mustard, and M.~Frank Morgan.
\newblock Revised {CRISM} spectral parameters and summary products based on the currently detected mineral diversity on {Mars}.
\newblock \emph{Journal of Geophysical Research: Planets}, 119\penalty0 (6):\penalty0 1403--1431, 2014.
\newblock ISSN 2169-9100.
\newblock \doi{10.1002/2014JE004627}.
\newblock URL \url{https://onlinelibrary.wiley.com/doi/abs/10.1002/2014JE004627}.
\newblock \_eprint: https://onlinelibrary.wiley.com/doi/pdf/10.1002/2014JE004627.

\bibitem[Zhang et~al.(2018)Zhang, Zuo, and Zhang]{zhang_ffdnet_2018}
Kai Zhang, Wangmeng Zuo, and Lei Zhang.
\newblock {FFDNet}: {Toward} a {Fast} and {Flexible} {Solution} for {CNN}-{Based} {Image} {Denoising}.
\newblock \emph{IEEE Transactions on Image Processing}, 27\penalty0 (9):\penalty0 4608--4622, September 2018.
\newblock ISSN 1941-0042.
\newblock \doi{10.1109/TIP.2018.2839891}.
\newblock URL \url{https://ieeexplore.ieee.org/abstract/document/8365806}.
\newblock Conference Name: IEEE Transactions on Image Processing.

\end{thebibliography}
\bibliographystyle{iclr2024_conference}

\newpage
\appendix
\section{Appendix}
\label{appendix}
\begin{figure}[!h]
    \centering
    \includegraphics[width=\textwidth]{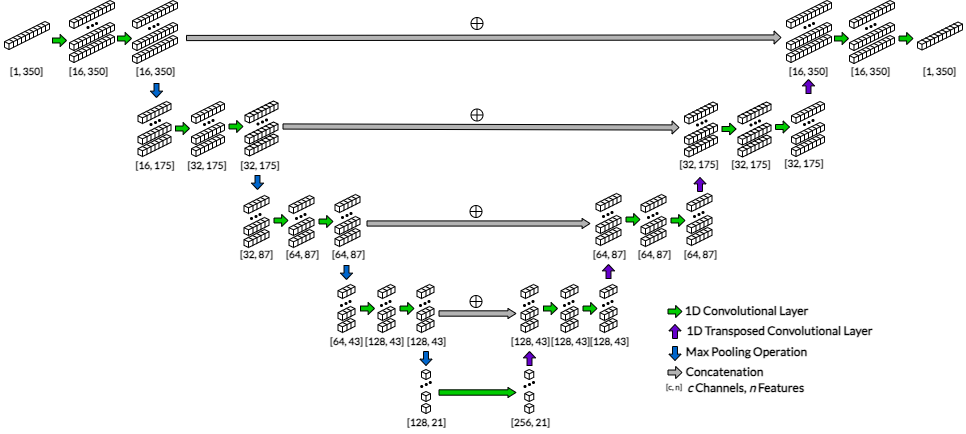}
    \caption{Noise2Noise 4 Mars (N2N4M) Network Architecture. All convolutional kernels were size 5.}
    \label{fig:Network_Architecture}
\end{figure}
Hyperparamter settings for N2N4M model training: learning rate = $1\times10^{-3}$, and batch size = 256. Model was trained for 100 epochs with early stopping implemented after 5 epochs of no decrease in validation score. Mean squared error (MSE) was used as the loss function, and ADAM as the optimiser. 

\end{document}